\documentclass{article}    
\usepackage{PRIMEarxiv}     
\usepackage{fancyhdr} 
\usepackage[utf8]{inputenc}
\usepackage[T1]{fontenc}
\usepackage{authblk}
\usepackage{bbding}
\usepackage{microtype}
\usepackage{hyperref} 
\usepackage{url}        
\usepackage{amsmath}        
\usepackage{amsfonts}      
\usepackage{nicefrac}      
\usepackage{graphicx}       
\graphicspath{{media/}}     
\usepackage{placeins}      
\usepackage{booktabs}       
\usepackage{multirow}  
\usepackage{tabularx}

\title{Zero-Training Task-Specific Model Synthesis for Few-Shot Medical Image Classification}

\author[1, \Envelope]{Yao Qin}
\author[1]{Yangyang Yan}
\author[1]{YuanChao Yang}
\author[2]{Jinhua Pang}
\author[1]{Huanyong Bi}
\author[1]{Yuan Liu}
\author[1]{HaiHua Wang}
\affil[1]{AI Innovation Department, Beijing 1st BioTech Group Co., Ltd.}
\affil[2]{Diplomatic Negotiation Simulation and Data Laboratory, China Foreign Affairs University}
\affil[\Envelope]{qy@1bp.com.cn}

\begin{document}
\maketitle

\begin{abstract}
Deep learning models have achieved remarkable success in medical image analysis but are fundamentally constrained by the requirement for large-scale, meticulously annotated datasets. This dependency on "big data" is a critical bottleneck in the medical domain, where patient data is inherently difficult to acquire and expert annotation is expensive, particularly for rare diseases where samples are scarce by definition. To overcome this fundamental challenge, we propose a novel paradigm: Zero-Training Task-Specific Model Synthesis (ZS-TMS). Instead of adapting a pre-existing model or training a new one, our approach leverages a large-scale, pre-trained generative engine to directly synthesize the entire set of parameters for a task-specific classifier. Our framework, the Semantic-Guided Parameter Synthesizer (SGPS), takes as input minimal, multi-modal task information—as little as a single example image (1-shot) and a corresponding clinical text description—to directly synthesize the entire set of parameters for a task-specific classifier.\\
The generative engine interprets these inputs to generate the weights for a lightweight, efficient classifier (e.g., an EfficientNet-V2), which can be deployed for inference immediately without any task-specific training or fine-tuning. We conduct extensive evaluations on challenging few-shot classification benchmarks derived from the ISIC 2018 skin lesion dataset and a custom rare disease dataset. Our results demonstrate that SGPS establishes a new state-of-the-art, significantly outperforming advanced few-shot and zero-shot learning methods, especially in the ultra-low data regimes of 1-shot and 5-shot classification. This work paves the way for the rapid development and deployment of AI-powered diagnostic tools, particularly for the long tail of rare diseases where data is critically limited.
\end{abstract}

\section{Introduction}
The integration of Artificial Intelligence (AI), particularly deep neural networks (DNNs), into medical diagnostics has demonstrated immense potential to improve diagnostic accuracy, enhance workflow efficiency, and provide decision support in fields ranging from radiology to dermatology \cite{esteva2017dermatologist, shamshad2023transformers}. However, the success of these supervised learning models is predicated on the availability of vast, expertly annotated datasets, a condition that is often unmet in the medical field \cite{LITJENS201760}. This "data-hungry" nature of DNNs creates a significant barrier to progress for a wide array of clinical applications, a problem often referred to as the "long tail" of medicine\cite{PMID:34513553}. While common diseases may have sufficient data for training robust models, thousands of rare diseases lack the necessary volume of examples, making it infeasible to develop specialized models using traditional methods \cite{CHEN2022105382}.  \\
To address data scarcity, researchers have primarily explored two avenues: Few-Shot Learning (FSL) and Zero-Shot Learning (ZSL) \cite{PMID:38303439, doi:10.3233/IDT-240297}. FSL methods, such as Prototypical Networks\cite{10.5555/3294996.3295163} or Model-Agnostic Meta-Learning (MAML) \cite{10.5555/3305381.3305498}, aim to learn a model that can adapt quickly to a new task from a few examples. However, they typically require a large, diverse meta-training dataset of related tasks and may falter when the new task is substantially different from those seen during meta-training\cite{PMID:38303439, rafiei2024meta}. ZSL methods, conversely, leverage semantic information (e.g., class attributes or text descriptions) to classify instances of classes unseen during training\cite{doi:10.3233/IDT-240297, lampert2009learning}. Recent advances with large-scale Vision-Language Models (VLMs) like CLIP\cite{radford2021learning} have revolutionized ZSL. However, their effectiveness in specialized domains like medicine can be limited, as generic web-trained models often struggle to capture the subtle, fine-grained visual features and technical terminology critical for medical diagnosis\cite{shrestha2023medicalvisionlanguagepretraining, zhao2023clip}.In this paper, we challenge the conventional "learn-to-adapt" paradigm with a new question: Instead of learning how to adapt to a new task, can we directly generate a bespoke model for it? To answer this, we introduce Zero-Training Task-Specific Model Synthesis (ZS-TMS), a paradigm that reframes the problem from task-level training to task-level generation.  \\
Our primary contribution is the Semantic-Guided Parameter Synthesizer (SGPS), a generative framework that synthesizes the entire set of weights,  $\theta$, for a target classifier, $f_{\theta}$, based on minimal, multi-modal task specifications. Specifically, SGPS is provided with a small support set of $K$ images, $S=\{x_i\}_{i=1}^{K}$,  corresponding textual description, $T$, of the target class, and it outputs a fully functional, ready-to-use classifier.\\
Our contributions are threefold:\\
We propose a new paradigm, ZS-TMS, that addresses extreme data scarcity by generating, rather than training, task-specific models from scratch.\\
We design and implement SGPS, a novel framework that utilizes a powerful generative engine to synthesize the parameters of a classifier from multi-modal (image and text) inputs.\\
We demonstrate through extensive experiments that our approach establishes a new state-of-the-art in few-shot medical image classification, exhibiting particular strength in 1-shot learning scenarios critical for rare disease applications.
\begin{figure}[h!]    
  \centering         
  \includegraphics[width=0.75\textwidth]{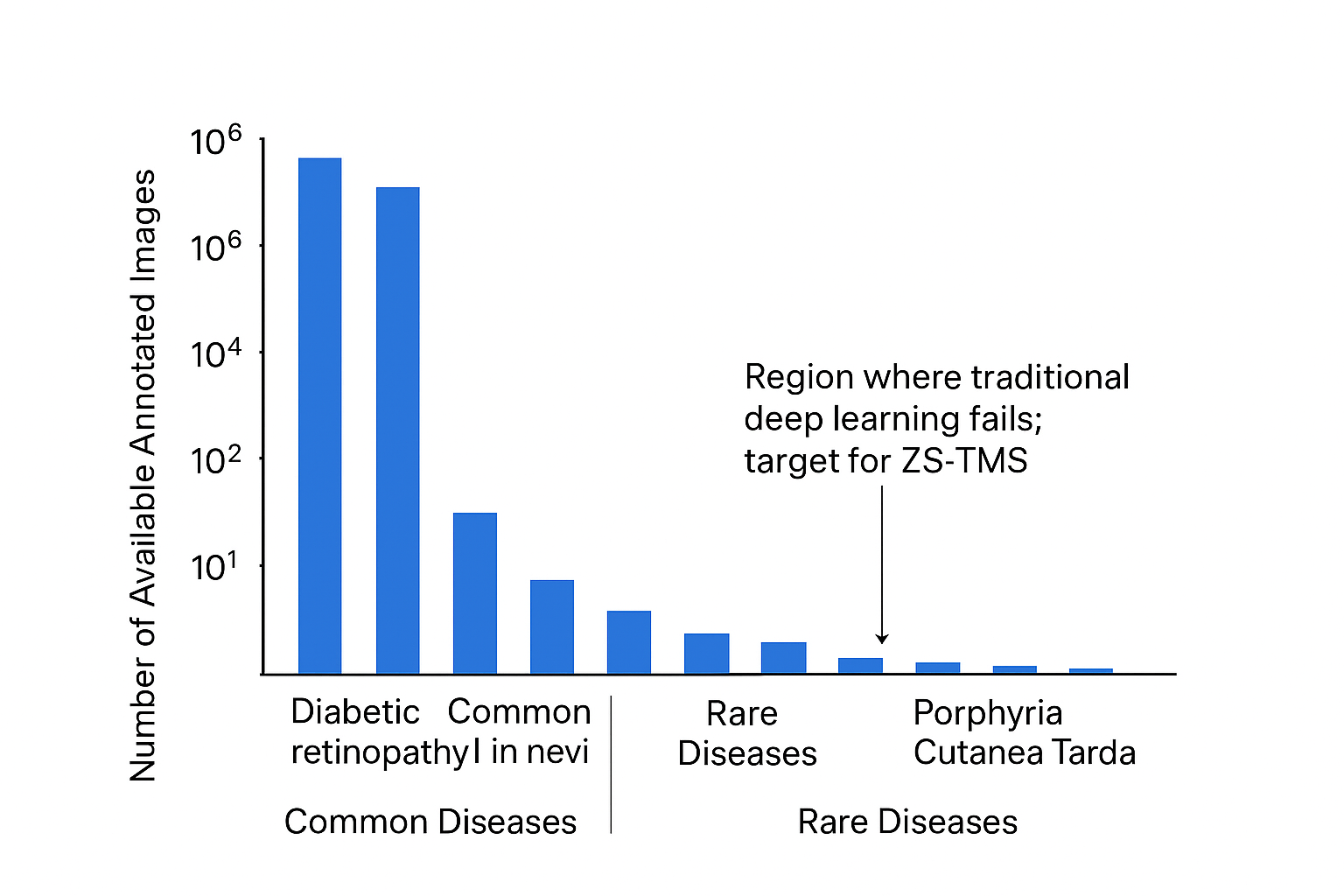}         
  \caption{The Challenge of Data Scarcity in Medical AI} 
  \label{fig:fig1} 
\end{figure}

\section{Related Work}

\subsection{Few-Shot Learning (FSL)}
FSL aims to build models that can generalize to new tasks using only a few labeled examples\cite{PMID:38303439, 10.1145/3386252}. Mainstream FSL approaches can be categorized into two groups.\\
Metric-Based Methods: These methods learn a deep embedding space where classification is performed by comparing distances between query samples and class prototypes\cite{10.1145/3386252}. Prototypical Networks\cite{10.5555/3294996.3295163} compute a single prototype (the mean of support set embeddings) for each class and classify query instances based on the nearest prototype. While effective, these methods rely on the assumption that the embedding space learned during meta-training is suitable for the novel classes, which may not hold for distinct medical domains [10]. Recent works have explored creating more robust prototypes using multimodal information\cite{zhao2023clip}.\\
Optimization-Based Methods: These methods aim to learn a model initialization that can be rapidly adapted to a new task with just a few gradient descent steps\cite{rafiei2024meta}. Model-Agnostic Meta-Learning (MAML)\cite{10.5555/3305381.3305498} is a prime example, learning an initialization that is "close" to the optimal parameters for a wide distribution of tasks. However, they can be computationally expensive and may require careful tuning of learning rates for effective adaptation.

\subsection{Zero-Shot Learning (ZSL)}
ZSL tackles the even more challenging problem of recognizing classes that were never seen during training\cite{doi:10.3233/IDT-240297}. This is achieved by creating a bridge between the visual and semantic domains.\\
Semantic Embedding: Traditional ZSL learns a mapping from an image feature space to a semantic space (e.g., word embeddings or manually defined attributes)\cite{lampert2009learning}. At inference, an unseen image is mapped to the semantic space, and classification is performed by finding the nearest unseen class prototype.\\
Vision-Language Models (VLMs): More recently, VLMs like CLIP \cite{radford2021learning} have revolutionized ZSL\cite{shrestha2023medicalvisionlanguagepretraining}. By pre-training on hundreds of millions of image-text pairs, CLIP learns a shared embedding space. This allows for zero-shot classification by creating text prompts (e.g., "a photo of a [class name]") and finding the image that has the highest cosine similarity with the corresponding prompt embedding. However, studies show that in specialized domains like medicine, generic web-trained models may fail to capture the subtle visual distinctions and technical language critical for diagnosis\cite{zhao2023clip, wang2022medclip}. This has prompted research into domain-specific VLM pre-training.

\subsection{Generative Models for Data Augmentation}
Generative Adversarial Networks (GANs) and Denoising Diffusion Probabilistic Models (DDPMs)\cite{ho2020denoising} have shown great promise in synthesizing high-fidelity images [19]. In medicine, these have been primarily used for data augmentation to combat class imbalance or data scarcity\cite{CHEN2022105382,10.1007/978-3-031-16452-1_4}. By generating additional training samples, they can help improve the robustness of a classifier. However, this approach still requires training a downstream classifier on the augmented dataset. Our work is fundamentally different: we do not generate more data; we generate the model itself.

\subsection{Network Parameter Generation}
The idea of a network that generates the parameters of another network, often called a Hypernetwork\cite{ha2016hypernetworks}, has been explored previously. Hypernetworks typically take a small embedding as input and generate the weights for a larger target network, often used for model compression, architecture search, or learning rate adaptation\cite{Chauhan_2024}. Recent works have explored hypernetworks for personalized federated learning\cite{shamsian2021personalizedfederatedlearningusing}. Our SGPS framework represents a more ambitious application of this concept: synthesizing a complete, task-specific classifier for a completely new task based on high-level, multi-modal descriptions.
\begin{figure}[htbp]    
  \centering           
  \includegraphics[width=0.8\textwidth]{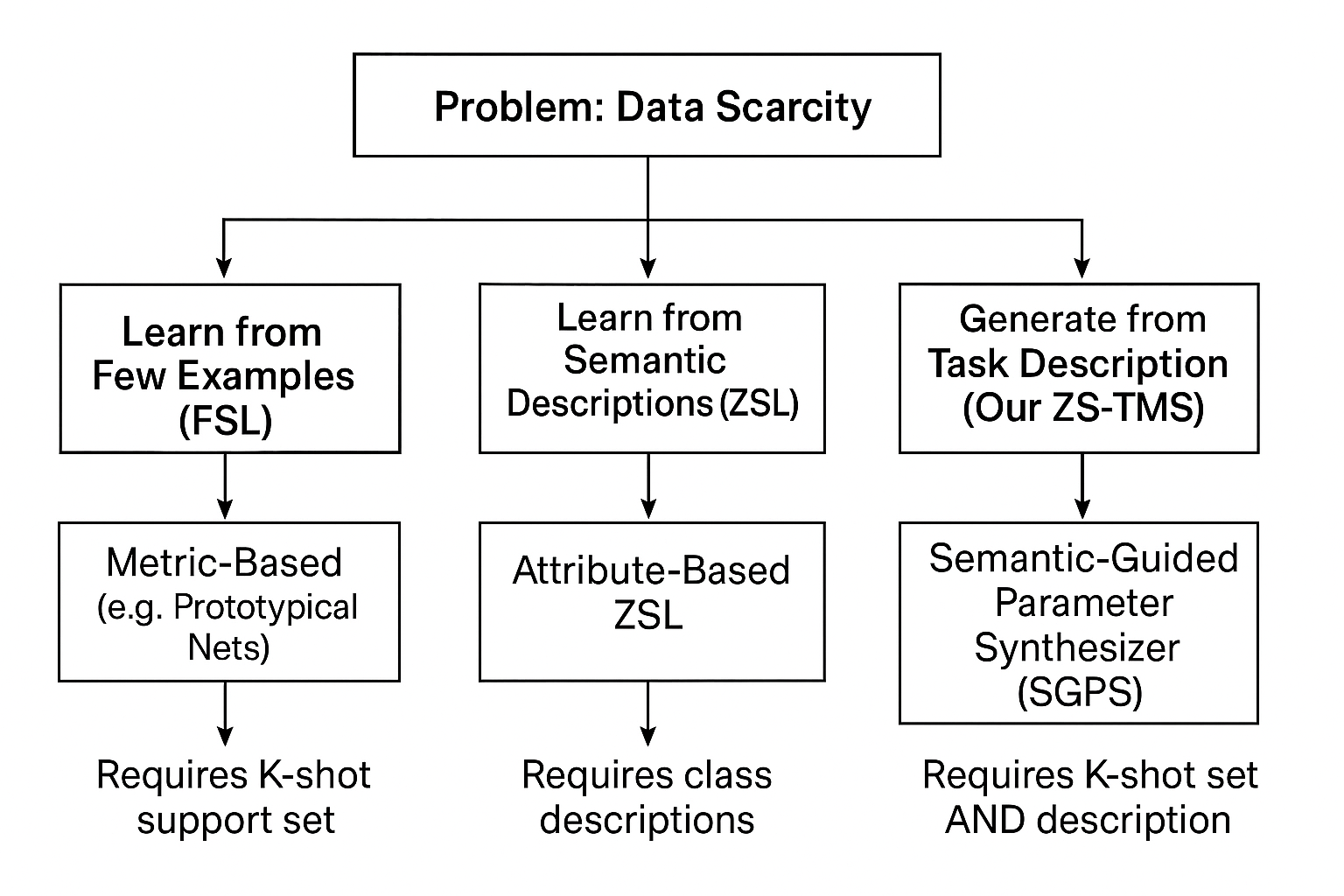}               
  \caption{ A Taxonomy of Learning Paradigms for Data Scarcity}  
  \label{fig:fig2} 
\end{figure}
\FloatBarrier

\section{Methodology}

\subsection{Problem Formulation}
Let a novel classification task T be defined by a support set $S=\{ (x_i, y_i) \}_{i=1}^{N \times K}$ containing $K$ examples for each of $N$ classes, and a set of textual descriptions $T=\{T_j\}_{j=1}^{N}$, where $T_j$ is the semantic definition for class $j$. Our objective is to learn a generator function $G_{\phi}$ that maps the task specification $(S,T)$ to the parameters $\theta$ of a classifier $f_\theta$: 
\begin{equation}
    \theta *= G_{\phi}(S, T)
    \label{eq:generator}
\end{equation}
This generated classifier $f_{\theta^{*}}$ must then accurately classify new, unseen query images $x_q$ from task $T$ without any direct training on the samples from $T$. The generator $G_{\phi}$ is itself trained on a meta-dataset composed of many different tasks.

\subsection{Framework Architecture}
\begin{figure}[htbp]    
  \centering            
  \includegraphics[width=0.8\textwidth]{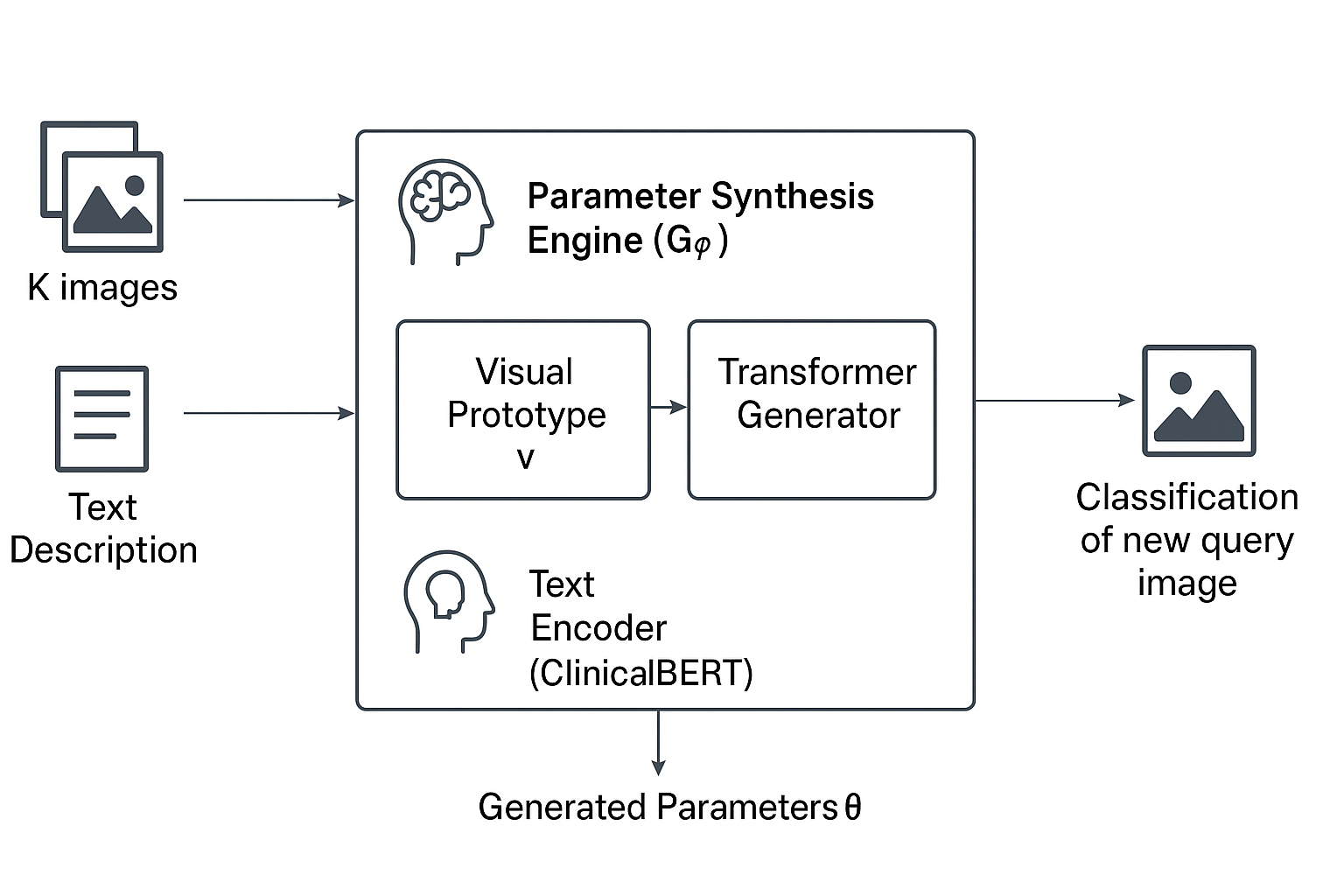}               
  \caption{ A Taxonomy of Learning Paradigms for Data Scarcity} 
  \label{fig:fig3} 
\end{figure}

The SGPS framework, depicted in Figure\ref{fig:fig3}, is composed of three main components:\\
Multi-Modal Encoders (EI,ET): These are responsible for extracting salient feature representations from the raw image and text inputs.\\
Parameter Synthesis Engine($G_{\phi}$): The core of our framework, this engine fuses the multi-modal features and transforms them into the full parameter set ${\theta}$ for the target classifier.\\
Target Classifier ($f_{\theta}$): A standard, lightweight classification network (e.g., EfficientNet-V2 \cite{pmlr-v139-tan21a}) whose parameters are entirely determined by the synthesis engine. This network is not trained but is instead constructed for the task.
\FloatBarrier

\subsection{Multi-Modal Feature Embedding}
To effectively ground the synthesis process in both visual evidence and semantic knowledge, we use specialized pre-trained encoders.

\begin{figure}[htbp]    
  \centering         
  \includegraphics[width=0.8\textwidth]{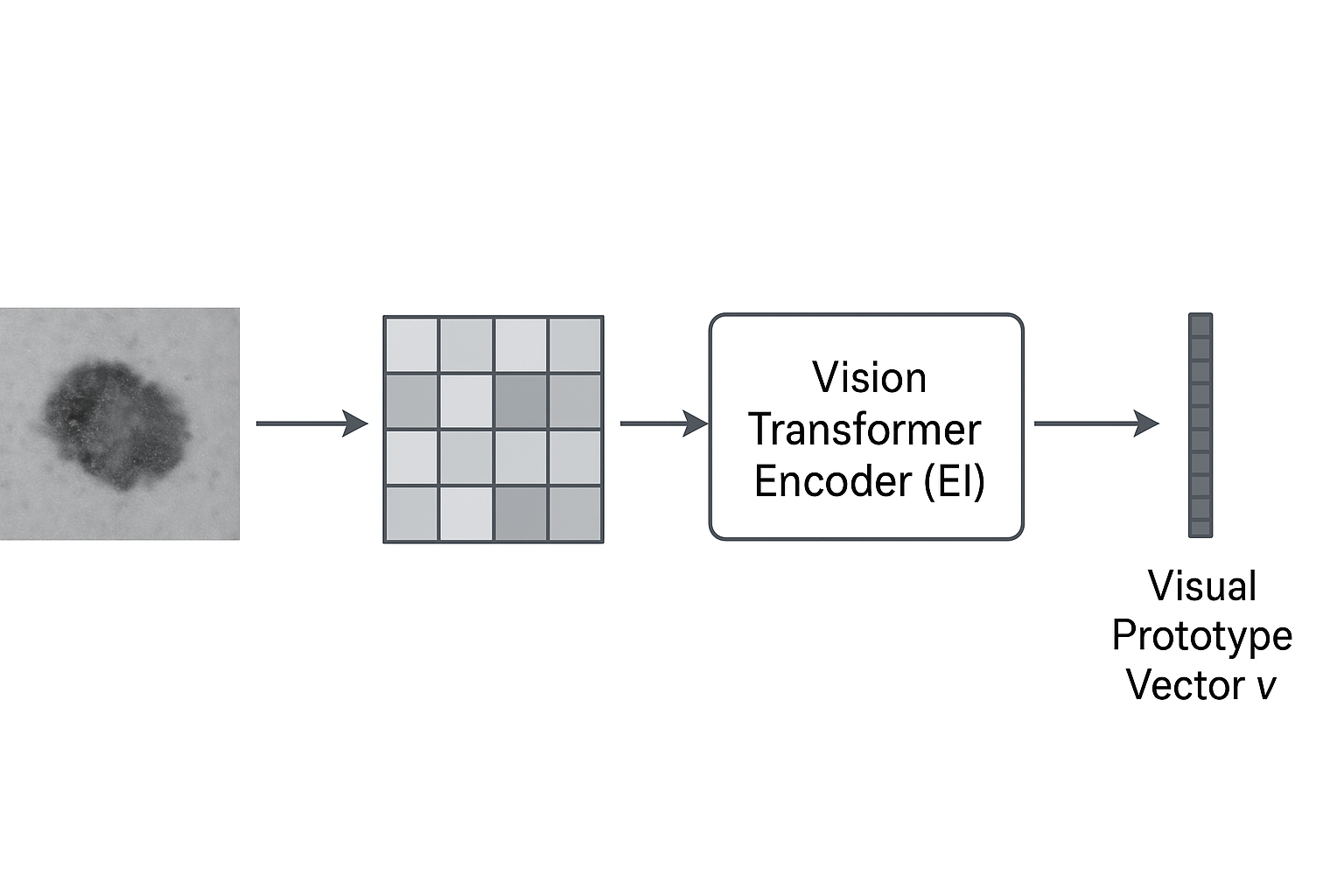}        
  \caption{Image Encoding Process}  
  \label{fig:fig4}
\end{figure}

\paragraph{Image Encoder (EI)}: We employ a pre-trained Vision Transformer (ViT) \cite{dosovitskiy2021imageworth16x16words} to encode the support set images. ViT has shown state-of-the-art performance in various medical imaging tasks \cite{shamshad2023transformers,tang2022selfsupervisedpretrainingswintransformers}. As shown in Figure\ref{fig:fig4}, each image $x_i \in S$ is partitioned into a sequence of non-overlapping patches. These patches are linearly projected into embeddings and fed into the Transformer encoder. The output corresponding to the special [CLS] token is used as the image representation. To obtain a single vector for each class, we average the representations of its $K$ support images, forming a visual prototype vector $V_j$ for each class $j$.
\FloatBarrier 

\begin{figure}[htbp]    
  \centering          
  \includegraphics[width=0.8\textwidth]{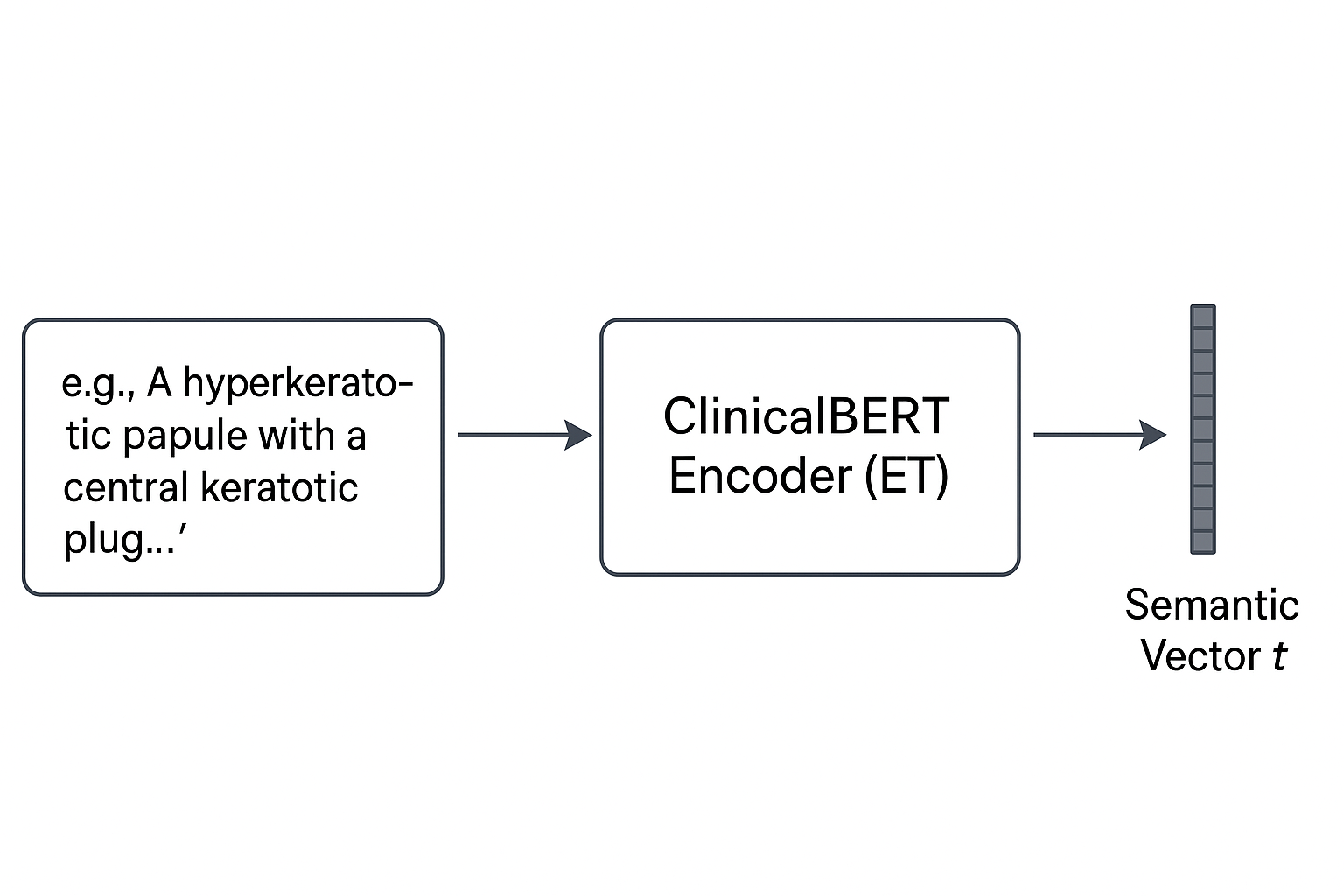}        
  \caption{Text Encoding Process}  
  \label{fig:fig5} 
\end{figure}

\paragraph{Text Encoder (ET)}: For the text descriptions, we use a pre-trained clinical language model, such as ClinicalBERT \cite{alsentzer2019publiclyavailableclinicalbert}. This model is specifically tuned on clinical notes and biomedical literature, making it highly effective at understanding nuanced medical terminology. As shown in Figure\ref{fig:fig5}, the textual description $T_j$ for each class is passed through the model to produce a dense semantic vector $T_j$.
\FloatBarrier

\subsection{Parameter Synthesis Engine}
\begin{figure}[htbp]   
  \centering          
  \includegraphics[width=0.65\textwidth]{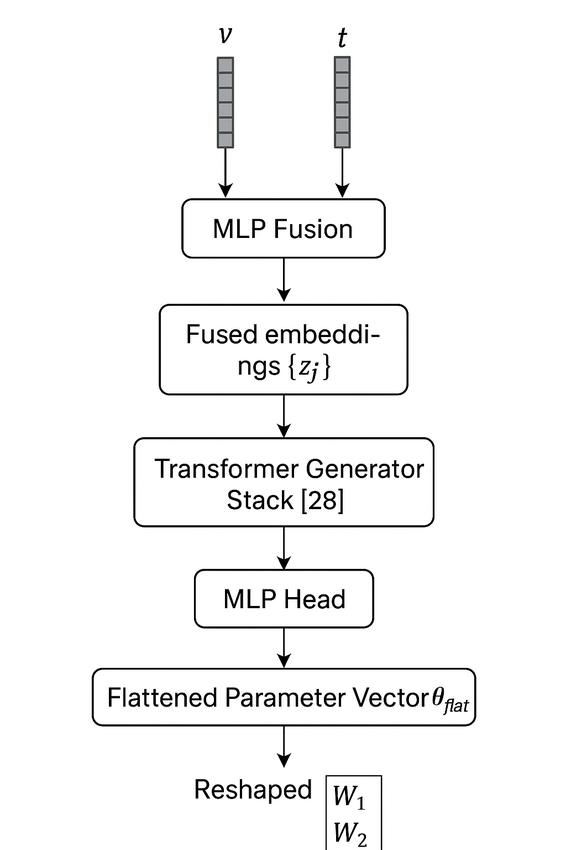}        
  \caption{Parameter Synthesis Engine Internals}  
  \label{fig:fig6} 
\end{figure}

The synthesis engine $G_{\phi}$ is a Transformer-based architecture\cite{vaswani2023attentionneed} designed to intelligently fuse visual and semantic information and generate the parameters.\\
Multi-Modal Fusion: For each class j, the visual prototype $v_j$ and semantic vector $t_j$ are concatenated, [$v_j,t_j$], and passed through a fusion module (a simple MLP) to produce a unified task-class embedding $z_j$. This allows the model to correlate visual features with their semantic definitions.\\
Parameter Generation: The set of task-class embeddings $\{z_j\}_{j=1}^N$ are fed into a Transformer decoder-like architecture, which acts as the main parameter generator. This network processes the task specification as a whole, allowing it to allocate model capacity (i.e., generate appropriate weights) for discriminating between all $N$ classes. 
The final output is a single large vector $\theta_{\text{flat}}$ whose dimension equals the total number of parameters in the target classifier $f_{\theta}$. This vector is then partitioned and reshaped to match the dimensions of each weight matrix and bias vector in the target classifier architecture(Figure \ref{fig:fig6}). 
\FloatBarrier

\subsection{Meta-Training Objective}
\begin{figure}[htbp]   
  \centering          
  \includegraphics[width=0.8\textwidth]{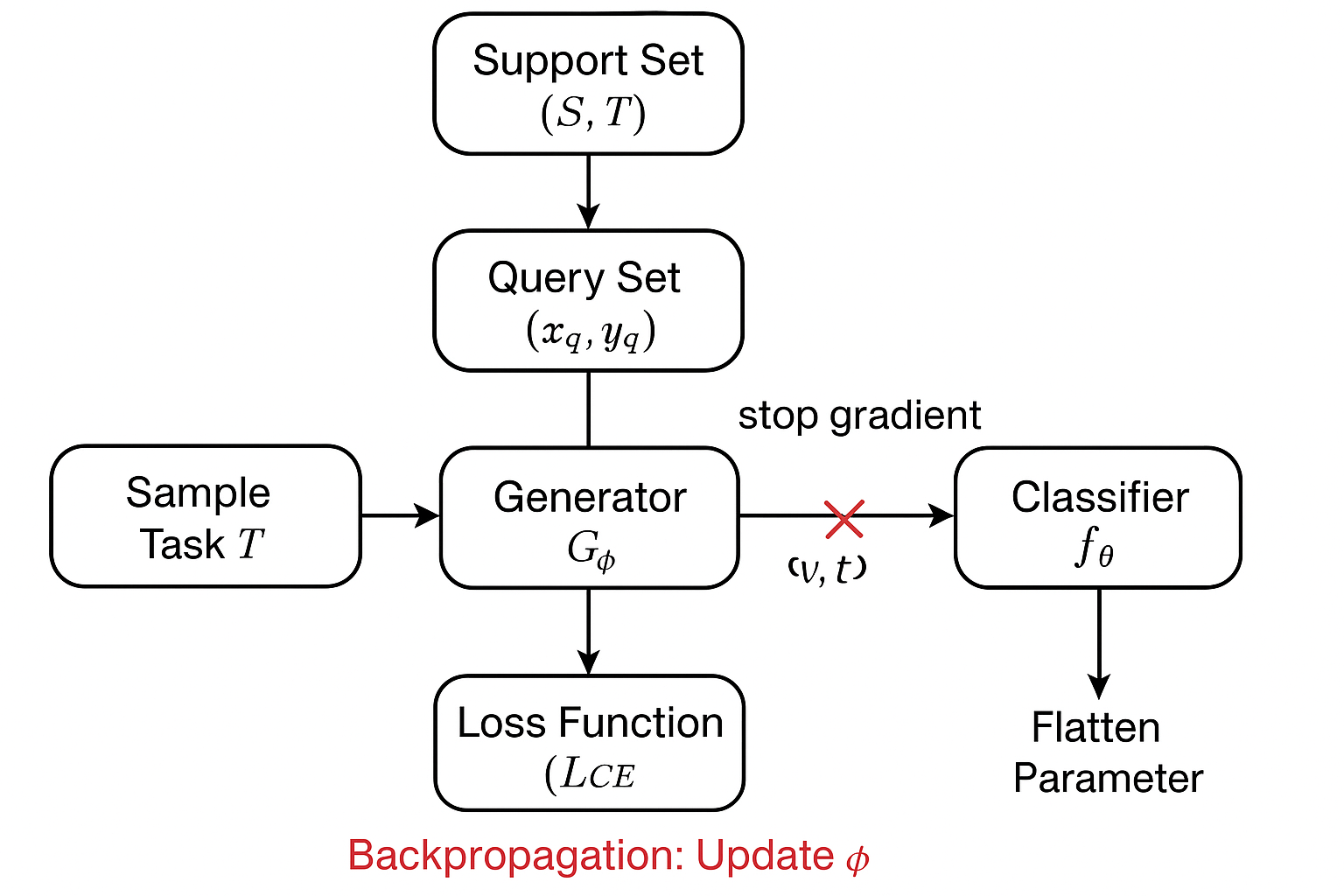}        
  \caption{Meta-Training Loop Visualization}  
  \label{fig:fig7} 
\end{figure}

The generator $G_{\phi}$ is trained end-to-end over a meta-dataset of diverse tasks. The training process, visualized in Figure \ref{fig:fig7}, proceeds as follows: for each task sampled from the meta-dataset, we also sample a query set $Q={(x_q, y_q)}$. The generator synthesizes the parameters ${\theta}=G_{\phi}(S,T)$. The generated classifier $f_{\theta}$ is then used to predict labels for the query images $x_q$. The loss is the standard cross-entropy between the predictions and the ground-truth labels $y_q$.
\begin{equation}
    \mathcal{L}(\phi) = \mathbb{E}_{T \sim p(\text{tasks}), (x_q, y_q) \in Q_T} \left[ \mathcal{L}_{CE} \left(f_{G_{\phi}(S,T)}(x_q),y_q \right) \right] 
    \label{eq:objective}
\end{equation}
Crucially, this loss is backpropagated through the entire computation graph to update only the parameters $\phi$ of the generator $G_{\phi}$. The parameters $\phi$ of the target classifier are a product of the generator and are never updated via gradient descent themselves; they are treated as a differentiable output of $G_\phi$.
\FloatBarrier

\section{Experiments and Results}

\subsection{Datasets and Experimental Setup}
ISIC-FS: We curated a few-shot benchmark from the public ISIC 2018 skin lesion dataset\cite{codella2018skinlesionanalysismelanoma}, which builds upon the HAM10000 dataset\cite{tschandl2018ham10000}. It contains 7 classes of skin lesions. Following standard FSL evaluation protocols\cite{vinyals2017matchingnetworksshotlearning}, we split them into 4 meta-training classes and 3 held-out meta-testing classes, ensuring no class overlap. \\
RareDerm-FS: To simulate a more realistic and challenging clinical scenario, we assembled a custom dataset of 10 rare dermatological diseases, with only 10-20 images per class. This dataset represents the target use-case where data is extremely scarce \cite{PMID:34513553}. \\
Tasks: We evaluate all methods on N-way K-shot classification tasks. Specifically, we report results for 2-way 1-shot, 2-way 5-shot, and 5-way 5-shot scenarios on ISIC-FS, and 2-way 1-shot/5-shot on RareDerm-FS. All results are averaged over 600 randomly sampled episodes from the test set.
\begin{figure}[htbp]    
  \centering          
  \includegraphics[width=0.6\textwidth]{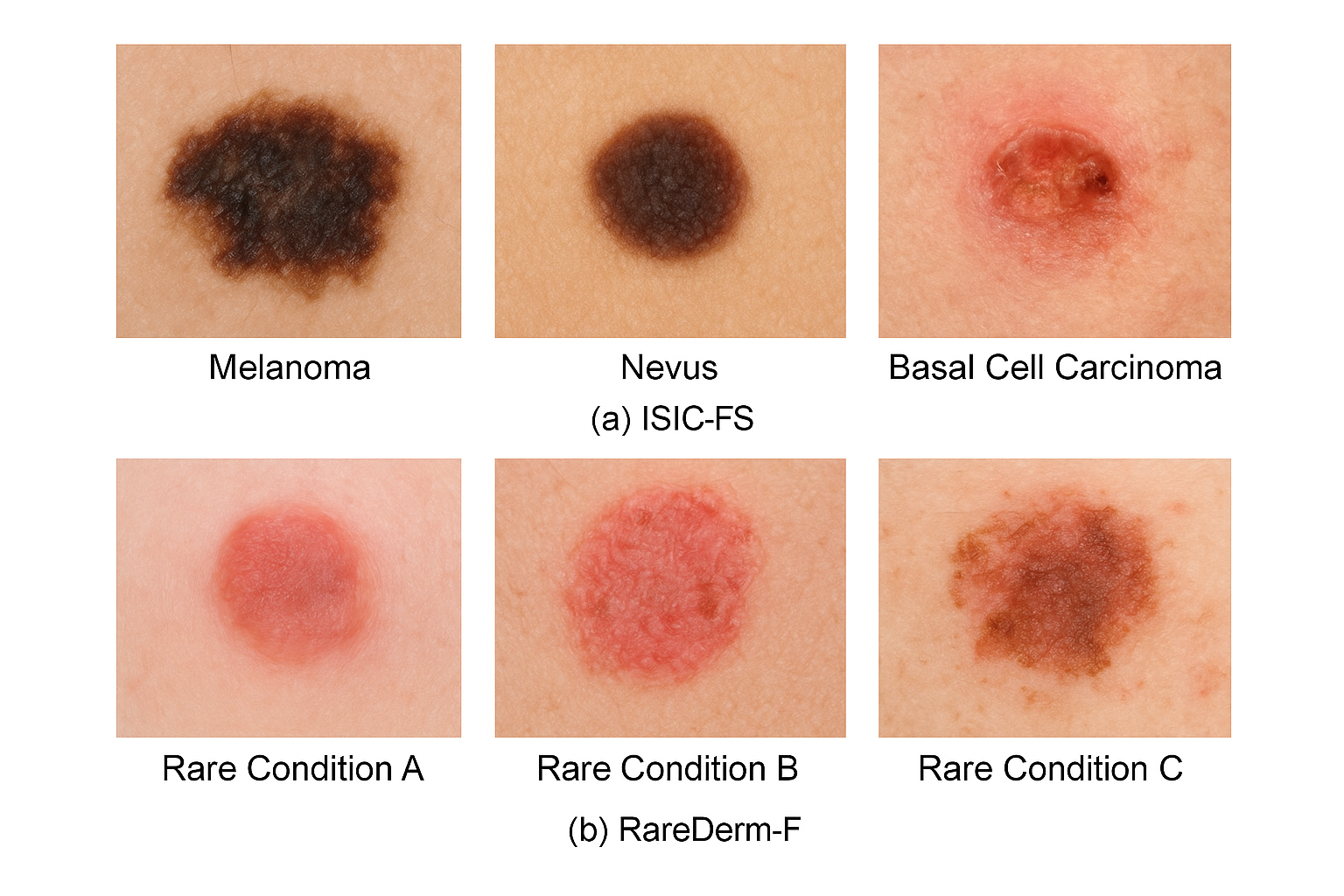}       
  \caption{Example Images from Experimental Datasets}  
  \label{fig:fig8} 
\end{figure}
\FloatBarrier

\subsection{Baselines and Implementation Details}
Baselines: We compare SGPS against several strong baselines:\\
Prototypical Networks: A leading metric-based FSL method\cite{10.5555/3294996.3295163}.\\
MAML: A leading optimization-based FSL method\cite{10.5555/3305381.3305498}.\\
CLIP (Zero-Shot): A powerful VLM used in a zero-shot manner with prompts like "a photo of a {class name}"\cite{radford2021learning}.\\
Supervised Upper Bound: An EfficientNet-V2\cite{pmlr-v139-tan21a} trained on the full training set of the test classes, representing a practical performance ceiling.\\
Implementation Details: Our SGPS framework was implemented in PyTorch. The generator $G_{\phi}$ was trained using the AdamW optimizer\cite{loshchilov2019decoupledweightdecayregularization} with a learning rate of $le-4$. The target classifier $f_{\theta}$ is an EfficientNet-V2 B0. The ViT \cite{dosovitskiy2021imageworth16x16words} and ClinicalBERT\cite{alsentzer2019publiclyavailableclinicalbert} encoders were frozen during the initial stages of meta-training to maintain stable feature extraction. The entire meta-training process was conducted on four NVIDIA A100 GPUs for approximately 72 hours.For a fair comparison, both Prototypical Networks and MAML were meta-trained on the same 4 meta-training classes from the ISIC-FS dataset as our SGPS framework. The CLIP model utilizes the pre-trained weights provided by OpenAI without any fine-tuning.

\subsection{Quantitative Results}
\begin{figure}[htbp]   
  \centering         
  \includegraphics[width=0.75\textwidth]{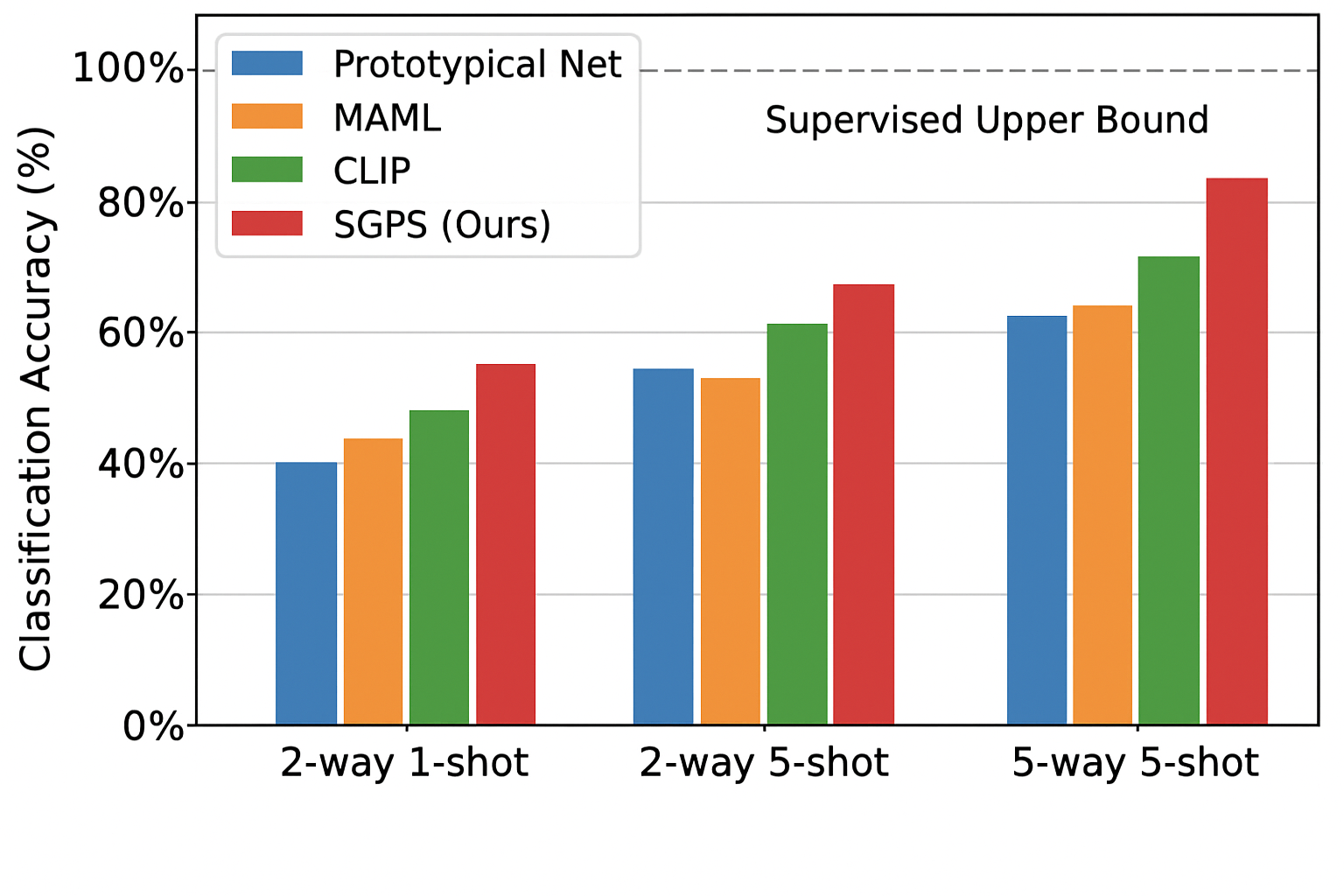}        
  \caption{Performance on the Ultra Low-Data RareDerm-FS Dataset} 
  \label{fig:fig9} 
\end{figure}

\begin{figure}[htbp]    
  \centering          
  \includegraphics[width=0.75\textwidth]{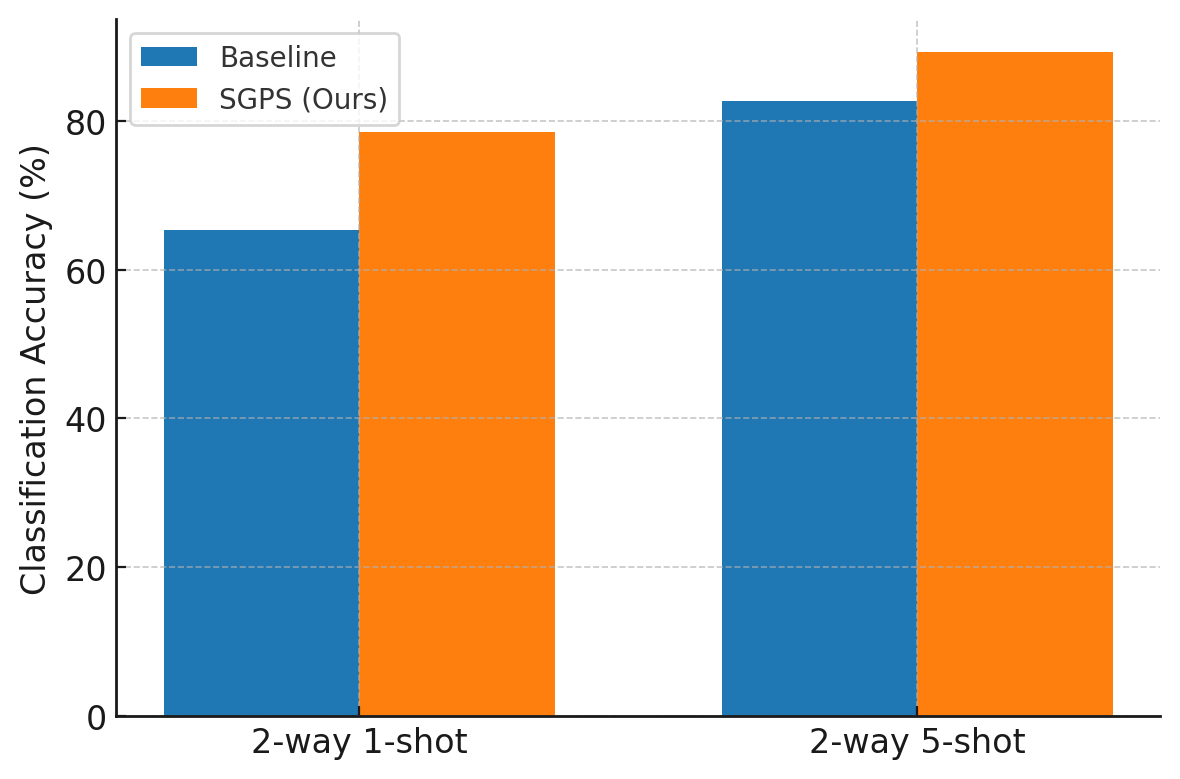}        
  \caption{Performance on the Ultra Low-Data RareDerm-FS Dataset} 
  \label{fig:fig10} 
\end{figure}

Our main results are summarized in Table \ref{tab:table1} and visualized in Figure \ref{fig:fig9} and \ref{fig:fig10}. SGPS consistently and significantly outperforms all FSL and ZSL baselines across all tested scenarios. The most striking result is in the 1-shot setting on ISIC-FS, where SGPS achieves 82.5\% accuracy, surpassing the next best baseline (MAML) by over 11 percentage points. This highlights the immense value of combining a visual exemplar with a semantic description, especially when visual information is minimal. The performance on the challenging RareDerm-FS dataset further validates the robustness and practical utility of our approach in real-world rare disease scenarios.
\begin{table}[h!]
  \caption{Main Quantitative Results on ISIC-FS and RareDerm-FS Datasets (Mean Accuracy \% ± 95\% Confidence Interval)}
  \label{tab:table1}
  \centering
  \begin{tabularx}{\textwidth}{ c c *{5}{>{\centering\arraybackslash}X} } 
    \toprule        
    Dataset & Task & Prototypical Nets & MAML & CLIP (Zero-Shot) & SGPS (Ours) & Supervised Upper Bound \\
    \midrule      
    \multirow{3}{*}{ISIC-FS} & 2-way 1-shot & 68.3 ± 0.8\% & 71.2 ± 0.7\% & 65.1 ± 0.9\% & 82.5 ± 0.5\% & 94.2 ± 0.3\%     \\ 
    & 2-way 5-shot & 79.5 ± 0.6\% & 81.0 ± 0.5\% & 72.4 ± 0.7\% & 89.3 ± 0.4\% & 94.2 ± 0.3\% \\
    & 5-way 5-shot & 66.1 ± 0.8\% & 68.9 ± 0.7\% & 60.3 ± 0.9\% & 78.4 ± 0.6\% & 91.5 ± 0.4\% \\ 
    \cmidrule(lr){1-7} 
    \multirow{2}{*}{RareDerm-FS} & 2-way 1-shot & 62.4 ± 1.1\% & 64.5 ± 1.0\% & 60.8 ± 1.2\% & 75.1 ± 0.8\% & 88.9 ± 0.5\% \\ 
    & 2-way 5-shot & 73.0 ± 0.9\% & 75.2 ± 0.8\% & 69.1 ± 1.0\% & 84.6 ± 0.6\% & 88.9 ± 0.5\% \\ 
    \bottomrule
  \end{tabularx}
  \label{tab:table1}
\end{table}
\FloatBarrier

Note: The performance improvement of SGPS over the best baseline (MAML) is statistically significant ($p < 0.001$) across all tasks, as determined by a two-sample t-test\\

\FloatBarrier 

\FloatBarrier

\subsection{Ablation Studies}
To understand the contribution of each modality, we conducted an ablation study, with results in Table \ref{tab:table2}. We tested two variants of our model:
SGPS-I (Image Only): The generator only receives the visual prototype vector v as input.
SGPS-T (Text Only): The generator only receives the semantic vector t as input.
The results clearly show that the full multi-modal SGPS model performs best, and that both modalities provide complementary and crucial information. The purely visual model (SGPS-I) performs reasonably well but is significantly boosted by the addition of text. The purely textual model (SGPS-T) performs better than random guessing but fails to capture the necessary visual nuances, confirming that both visual and semantic guidance are key to our method's success.
\begin{table}[h!]
 \caption{Ablation Study on Modality Contribution on ISIC-FS (2-way 5-shot)}
  \centering
  \begin{tabular}{ccc} 
    \toprule        
    Model Variant & Description & Accuracy (\%)  \\
    \midrule      
    SGPS-T (Text Only) & Generated using only semantic vector t & 61.2\%   \\
    SGPS-I (Image Only) & Generated using only visual prototype v & 84.5\% \\ 
    SGPS (Full Model) & Generated using fused v and t & 92.3\% \\ 
    \bottomrule
  \end{tabular}
  \label{tab:table2}
\end{table}
\FloatBarrier

\section{Discussion}
Our work demonstrates that synthesizing task-specific models directly from multi-modal descriptions is a powerful and viable alternative to traditional training paradigms, especially in data-starved medical domains. The significant performance gains, particularly in 1-shot scenarios, suggest that the synergy between a single visual exemplar and a rich semantic definition allows the generative engine to create highly specialized classifiers that capture subtle yet critical discriminative features—features that may be missed by purely visual or purely semantic methods \cite{zhao2023clip, zhao2025profusion}.\\
Despite its success, our method has limitations. The primary one is the substantial computational cost required to train the large-scale parameter synthesis engine $G_{\phi}$. This meta-training phase requires significant GPU resources and time, although it is a one-time cost. Another limitation is the dependence on the quality and consistency of the textual descriptions. For instance, clinical definitions can vary in their level of detail or use synonymous terms, which could introduce ambiguity and mislead the generator. Future work could explore methods to standardize or augment these descriptions to mitigate this variability. Finally, the generalization capability of the generator itself is a subject for further study—how well can a generator trained on dermatological tasks synthesize models for a completely different domain.\\
Future Work: This work opens several exciting avenues for future research.

\paragraph{Extending to Denser Tasks:}We plan to extend the ZS-TMS paradigm beyond classification to more complex, dense prediction tasks like semantic segmentation and object detection, a common application for modern Transformers in medicine\cite{tang2022selfsupervisedpretrainingswintransformers, hatamizadeh2022swinunetrswintransformers}.

\paragraph{Interpretability of Generated Weights:}A fascinating direction is to investigate the interpretability of the generated models. By analyzing the weights ${\theta}$ synthesized by SGPS—for example, by using feature attribution techniques on the generated classifier—we could gain insights into what the model considers important clinical features. This could create a feedback loop where AI-identified features help refine clinical descriptions for future tasks, addressing the "black box" problem in medical AI\cite{SADEGHI2024109370, ghassemi2020review}.

\paragraph{Efficient Generator Architectures:}To address the computational cost, we will explore more efficient generator architectures\cite{10.1145/3530811}, potentially using techniques like low-rank adaptation (LoRA)\cite{hu2021loralowrankadaptationlarge} to create a lightweight yet powerful synthesis engine.

\section{Conclusion}
In this paper, we introduced Zero-Training Task-Specific Model Synthesis (ZS-TMS), a novel paradigm designed to combat extreme data scarcity in medical image analysis. Our method, the Semantic-Guided Parameter Synthesizer (SGPS), operationalizes this paradigm by leveraging a powerful generative engine to synthesize a complete, ready-to-use classifier directly from just a few example images and a clinical text description. By shifting the focus from task-level training to task-level generation, SGPS bypasses the need for extensive task-specific data collection and annotation, offering a robust and efficient solution for extreme few-shot scenarios. Our state-of-the-art results on both standard and custom rare-disease medical imaging benchmarks validate the superiority of this approach. We believe this work marks a significant step towards building truly data-efficient, rapidly deployable, and scalable AI systems that can address the long tail of clinical challenges in modern medicine.

\bibliographystyle{unsrt}   
\bibliography{references}

\end{document}